\def\year{2022}\relax
\documentclass[letterpaper]{article}
\pdfoutput=1

\usepackage{titling}
\usepackage{aaai22} 
\usepackage{times} 
\usepackage{helvet} 
\usepackage{courier} 
\usepackage[hyphens]{url} 
\usepackage{graphicx} 
\usepackage{graphicx} 
\usepackage{natbib} 
\usepackage{caption} 
\DeclareCaptionStyle{ruled}%
{labelfont=normalfont,labelsep=colon,strut=off}
\usepackage[utf8]{inputenc} 
\usepackage[T1]{fontenc}    
\usepackage{hyperref}       
\usepackage{url}            
\usepackage{booktabs}       
\usepackage{amsfonts}       
\usepackage{nicefrac}       
\usepackage{microtype}      
\usepackage{xcolor}         
\usepackage{amsmath}
\usepackage{mathtools}
\usepackage{pdflscape}
\usepackage{color}
\usepackage{array}
\usepackage{subcaption}
\usepackage[title]{appendix}
\usepackage{algorithm}
\usepackage{algorithmic}

\newcommand{\matr}[1]{\mathbf{#1}}
\newcommand{\vect}[1]{\boldsymbol{#1}}
\graphicspath{ {./image/} }
\title{Transformers as Policies for Variable Action Environments}
\author{
Niklas Zwingenberger\\
}
\affiliations {
University of California, Los Angeles\\
Los Angeles, CA 90095 \\
\texttt{niklasz@ucla.edu} \\
}

\begin{document}
\maketitle
\begin{abstract}

In this project we demonstrate the effectiveness of the transformer encoder as a viable architecture for policies in variable action environments. Using it, we train an agent using Proximal Policy Optimisation (PPO) on multiple maps against scripted opponents in the \emph{Gym-$\mu$RTS} environment. The final agent is able to achieve a higher return using half the computational resources of the next-best RL agent \cite{DBLP:journals/corr/abs-2105-13807} which used the \emph{GridNet} architecture \cite{Han2019GridWiseCF}.

\end{abstract}

\section{Introduction}
In the past years significant progress has been made in variable action environments, particularly in Real Time Strategy games (RTS). For example the agent AlphaStar\cite{alphastarblog} has achieved a grand-master level in Starcraft II, consistently defeating human professional players. However, such state-of-the-art agents require months of training on powerful machines equipped with hundreds of CPUs and TPUs, which makes them impractical for research. In response to this, Huang \emph{et al.} have developed a more accessible environment known as \emph{Gym-$\mu$RTS}\cite{DBLP:journals/corr/abs-2105-13807}, suitable for training on a single commercial machine.

Huang \emph{et al.} also demonstrated 2 state-of-the-art methods to train in the environment \emph{Gym-$\mu$RTS}. These are \emph{Unit Acton Simulation} (UAS) \cite{DBLP:journals/corr/abs-2105-13807} and \emph{GridNet} \cite{Han2019GridWiseCF} respectively. However, we found that both methods suffer from limitations, namely that UAS requires access to the environment's model and that \emph{GridNet} does not computationally scale well with the size of the environment. Our new approach using a feature map and transformer architecture mitigates these issues.

In this report, we begin by discussing the relevant background in section \ref{sec:background}, with the description of the architecture in section \ref{sec:method} and relevant results in section \ref{sec:results}. The agent's implementation, trained model and recordings are made available in Appendix \ref{app:source_code}.

\section{Background}\label{sec:background}

\subsection{Variable Action Environments}
We can define a variable action environment as a tuple $(S,A,P,\rho_0,\gamma, r, T)$ where $S$ is the state-space, $A$ is a discrete action space, $P:S\times A\times S \rightarrow [0,1]$ is the transition probability, $\rho_0 : S\rightarrow [0,1]$ is the initial state distribution, $\gamma \in [0,1]$  is the discount factor, $r: S\times A \rightarrow \mathbb{R}$ is the reward function and $T$ is the maximum episode length. The variable aspect can be introduced by letting our action space $A=U_1\times U_2 ... \times U_k$ be expressed as a Cartesian product of sub-actions $U_i, i\in {1...k}$ where $k:S \rightarrow \mathbb{N}^+$ is the number of sub-actions that needs to be taken, dependent on the current state.

It should noted that while $k$ can in theory be arbitrarily large, it in most practical cases is bounded by some maximum number of sub-actions $N$. For example, in Starcraft II, if we consider each unit $i$ to take some sub-action from the space $U_i$, then $N$ must be equal to the maximum number of units the game can support (accounts suggest that this ranges from 1700-6400, depending on machine specs)\cite{starcraft_max_units}. For our project, we can show that \emph{Gym-$\mu$RTS} also has a clear upper bound $N$, which in practice is never reached and our architecture takes advantage of this fact.

\subsection{Transformers}
The transformer architecture is a neural network that uses self-attention, originally introduced for the task of machine translation \cite{DBLP:journals/corr/VaswaniSPUJGKP17}. Since then it has seen a wide range of applications in NLP tasks \cite{DBLP:journals/corr/abs-1810-04805}, vision tasks \cite{vision_transformers}, reinforcement learning tasks \cite{BRAMLAGE202210} and multi-modal tasks \cite{DBLP:journals/corr/abs-1908-02265}. In reinforcement learning in particular, decision transformers \cite{DBLP:journals/corr/abs-2106-01345} have been used for their sequence modelling properties to predict actions across entire trajectories, which is useful for Non-Markovian environments. 

In the context of this project it is sufficient to understand the transformer layer and \textit{multi-head self-attention} as presented by \cite{DBLP:journals/corr/VaswaniSPUJGKP17}:
\[  
\begin{split}
\text{Attention}(\matr{Q},\matr{K}, \matr{V}) = \text{softmax}\left(\frac{\matr{Q}\matr{K}^T}{\sqrt{d_k}} \matr{V}\right), \\
\text{MultiHead}(\matr{Q},\matr{K}, \matr{V}) = \text{Concat}\left(\text{head}_1,...,\text{head}_h\right)\matr{W}^O, \\
\text{head}_i = \text{Attention}(\matr{Q}\matr{W}_i^Q,\matr{K}\matr{W}_i^K, \matr{V}\matr{W}_i^V)
\end{split}
\]

\subsection{Proximal Policy Optimisation}
Policy gradient methods are a popular choice when training agents with large action spaces as they optimise the policy $\pi_\theta$ directly, which usually allows for a more targeted exploration of promising actions. The optimisation works by performing a gradient ascent on an objective function $J(\theta)$, where $\theta$ are the model parameters that we optimise. We commonly define $J$ to be the expected discounted return of a trajectory:

\[  
\begin{split}
J(\theta) 
= \mathbb{E}_{\tau \sim \pi_\theta, \rho_0}[G_0] 
= \mathbb{E}_{\tau \sim \pi_\theta, \rho_0}\left[\sum_{t=0}^{T-1}\gamma^t r_t\right], \\
\tau=(s_0,a_0,r_0,...s_{T-1},a_{T-1},r_{T-1})
\end{split}
\]

Where $\tau$ is the trajectory obtained from sampling the initial distribution $\rho_0$ and parameterised policy $\pi_\theta$. We maximise the expectation by taking the \textit{policy gradient} $\nabla_\theta J(\theta)$ and updating the parameters using $\theta := \theta + \alpha\nabla_\theta J(\theta)$, where $\alpha\in(0,1]$ is the learning rate. The gradient itself is given by \cite{sutton_barto_2020}:
\[  
\nabla_\theta J(\theta) 
= \mathbb{E}_{\tau \sim \pi_\theta, \rho_0}\left[\sum_{t=0}^{T-1}\nabla_\theta \log{\pi_\theta(a_t|s_t) G_t} \right] 
\]

As we are sampling entire trajectories, this gradient is subject to high variance\cite{sutton_barto_2020} and over the past years, many efforts have been made to reduce it \cite{DBLP:journals/corr/SchulmanLMJA15}. A particular enhancement of the above is \textit{Proximal Policy Optimisation} (PPO)\cite{https://doi.org/10.48550/arxiv.1707.06347}, with the following objective function:

\[  
\begin{split}
J(\theta)
= \mathbb{E}\left[\min(r_t(\theta)\hat{A}_t,  \text{clip}(r_t(\theta), 1 - \epsilon, 1 + \epsilon)\hat{A}_t\right], \\
r_t(\theta) = \frac{\pi_\theta(a|s)}{\pi_{\theta_{old}}(a|s)}, \\
\hat{A}_t = \sum_{l=0}^{T-1} (\gamma\lambda)^l(r_t + \gamma V_\theta(s_{t+l+1}) - \gamma V_\theta(s_{t+l}))
\end{split}
\]

PPO is an actor-critic algorithm, meaning it seeks to optimise both a parameterised policy $\pi_\theta(a|s)$ and a value estimate $V_\theta(s)$. The objective function limits the scope of gradient updates based on the advantage $\hat{A}_t$. In this case, it is the Generalised Advantage Estimate (GAE) \cite{schulman_high_dimensional_2018}.

We have re-used  Huang \emph{et al.}'s implementation of PPO \cite{PPO_blog}\cite{DBLP:journals/corr/abs-2005-12729}\cite{DBLP:journals/corr/abs-2006-05990} as we would like to draw comparisons between our results, focusing on the neural network implementations underlying $\pi_\theta(a|s)$ and $V_\theta(s)$. 

\subsection{$\mu$RTS Environment}
\begin{figure}
\centering
\includegraphics[scale= 0.088]{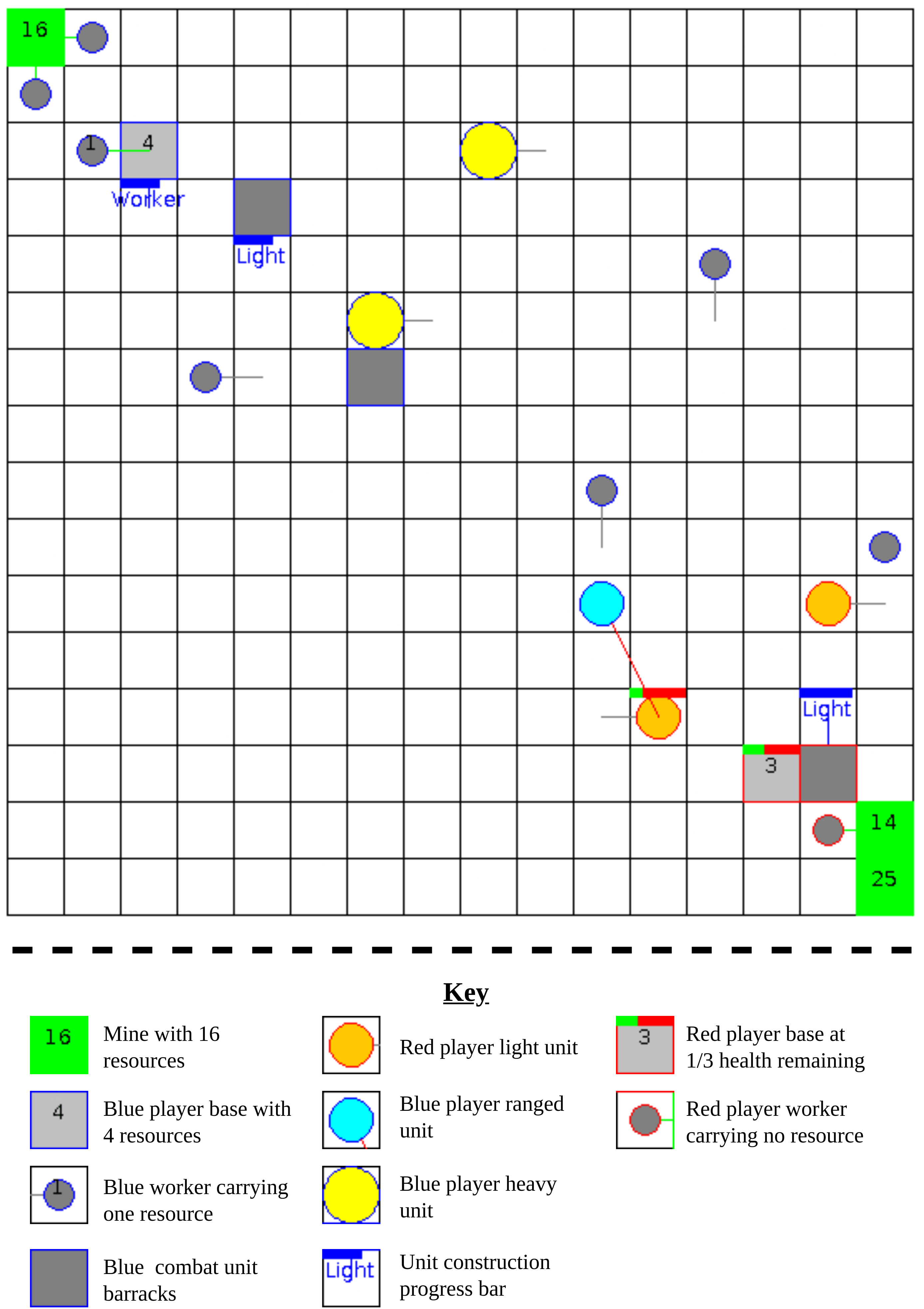}
\caption{Example of elements of a $\mu$RTS environment. In the above, there is a blue player based in the top-left corner of the map and a red player based in the bottom right. Both players are using workers to harvest resources from mines and are currently fighting using combat units. Every unit can only move to free horizontally or vertically adjacent cells and every action (including moving) takes several time steps to complete.}
\label{fig:example_grid}
\end{figure}

The environment is a grid-world that supports 1-4 players, where each one controls units which must gather resources and build structures and more units to defeat each other in combat. An episode begins with each player's units starting in a mirrored section of a map and ends when only one player's units remain or some step limit is reached. This environment can be configured to use different map sizes, layouts, partial observability, stochasticity and more. In our project we will be using 1 agent to interact with scripted AIs in the \texttt{basesWorkers8x8.xml} and \texttt{basesWorkers16x16.xml} maps. Both maps are 2 player and fully observable, differing only in size ($8\times8$ vs $16\times16$) and available resources (50 vs. 100). Please see  Figure \ref{fig:example_grid} for an example state of the $16\times16$ map.

\subsubsection{Observation Space}
During every step $t$ in an episode we receive an observation tensor $s_t$ of shape $(height \times width \times features)$. The first two dimensions describe a particular cell in the grid-world and the features dimension describes its state. For example, using the $8\times8$ map, it has the shape $(8\times 8 \times 27)$, where the 27 features describe 5 pieces of information: hit points (HP), resources, the owning player, unit type and current action.  Each piece of information is expressed using a one-hot encoding, where for example we can express a cell having 2 HP as $[0,0,1,0,0]$ or having $\geq 4$ resources as $[0,0,0,0,1]$. For a full description see the upper section of Table \ref{fig:action_obs_spaces} with further examples in Appendix \ref{app:obs_example}. 

\begin{figure*}
\centering
\includegraphics[scale= 0.20]{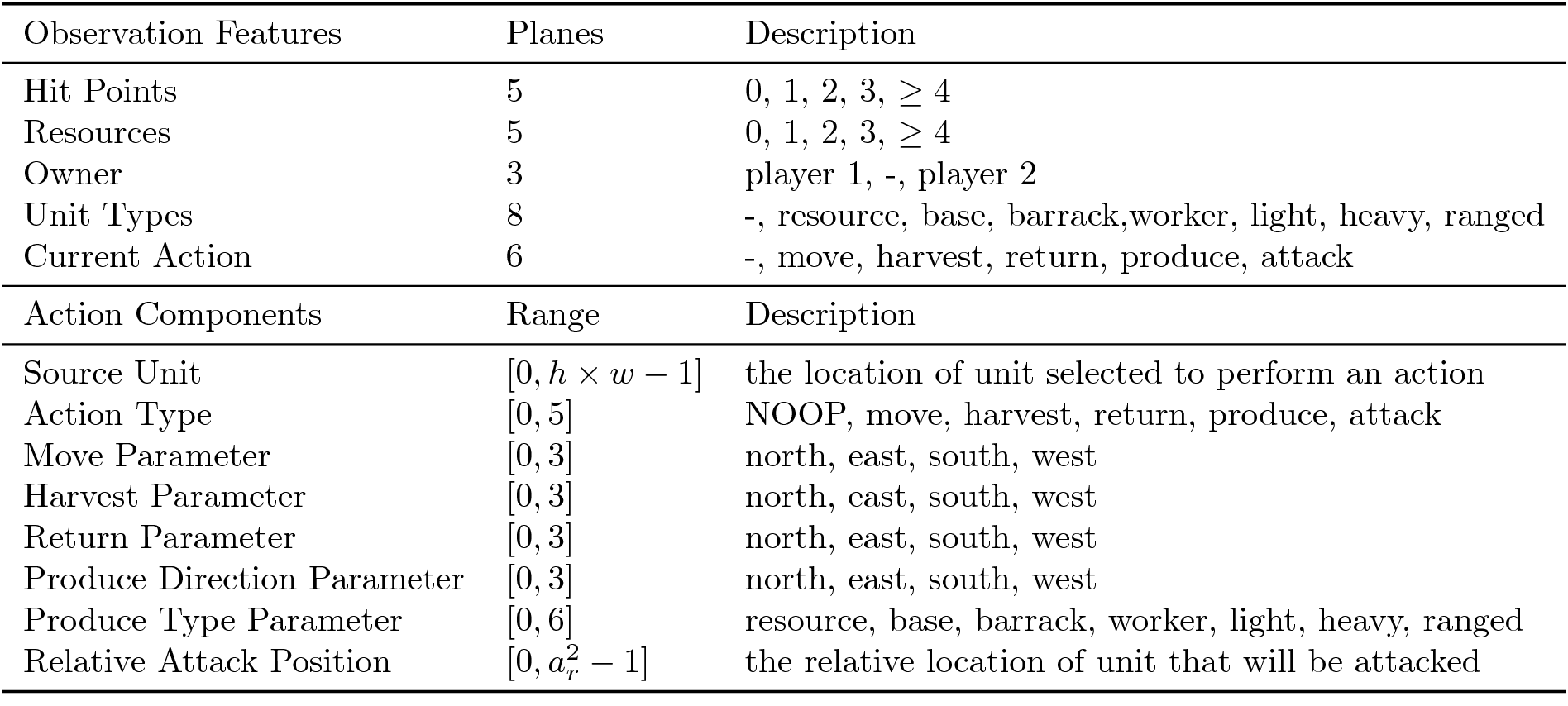}
\caption{Description observation and action space in $\mu$RTS environment \cite{DBLP:journals/corr/abs-2105-13807}. $a_r$ is the maximum attack range and is usually set to 7.}
\label{fig:action_obs_spaces}
\end{figure*}

\subsubsection{Action Space}
From the above we know that the observation restricts each cell to have at most 1 unit on it, which limits how many units can be on the map at any time. Each cell $(x,y)$ with a unit can take some sub-action $U_{x,y}$, allowing at most $N=height\times width$ sub-actions to be taken per time step. The components and possible values of an action are shown in the lower section of Table \ref{fig:action_obs_spaces}. Using them we can show that for some cell $(x,y)$ there are $6\times 4\times 4\times 4\times 4\times 7\times 7^2 = 526848$ possible discrete values that a sub-action can assume. This is quite large number and having a policy network generate half a million logits for every grid cell is too expensive, especially when taking the gradient afterwards.

Instead we make an independence assumption between each action component \cite{DBLP:journals/corr/abs-2105-13807} and define the probability of our policy $\pi_{\theta,x,y}(u_t|s_t)$ choosing a sub-action $u_t\in U_{x,y}$ given some $s_t\in S$ as:

\[ 
\begin{split}
\pi_{\theta}(a_t|s_t) = \prod_{x\in height} \prod_{y\in width} \pi_{\theta,x,y}(u_t|s_t), \\
\pi_{\theta,x,y}(u_t|s_t)= \prod_{u^d_t\in D} \pi_\theta(u^d_t |s_t), \\
D=\{u_t^{\text{Source Unit}}, u_t^{\text{Action Type}}, u_t^{\text{Move Parameter}}, u_t^{\text{Harvest Parameter}}, \\
u_t^{\text{Return Parameter}}, u_t^{\text{Produce Direction Parameter}}, u_t^{\text{Produce Type Parameter}}, \\
u_t^{\text{Relative Attack Position}}\}
\end{split}
\]

where $\pi_{\theta}$ is the joint policy of each individual policy $\pi_{\theta,x,y}$ per grid cell and $D$ is the set of action components. This way only $6 + 4 + 4 + 4 + 4 + 7 + 7^2 = 78$ logits are needed per sub-action.

Additionally, the \emph{Gym-$\mu$RTS} also provides built-in \textit{invalid action masking} \cite{huang2020closer}, which replaces the logits for nonsensical actions with high negative values. Then when we take the softmax over a distribution parameterised by these logits, the probability of such actions is 0. As an example, light, heavy and ranged combat units cannot produce any buildings or units, so the masking sets the probability of  the "Action Type Parameter" action component assuming value "produce" to 0.

\subsection{GridNet}
The \emph{GridNet} approach is a policy that $\pi_{\theta,x,y}(a_t|s_t)$ yields a sub-action $u_i$ for every grid cell $i\in\{1,2,..., height \times width\}$. This approach is convenient as it is quite easy to implement a policy that takes a fixed-size input (e.g the grid world) and yields a fixed-size output (action logits for every cell). The \emph{GridNet} itself uses a convolutional neural net (CNN) for this. A problem is that in practice for \emph{Gym-$\mu$RTS} and most other RTS games, controllable units do not exist on every cell, so a large chunk of the sub-actions are computed only to be thrown away. This computational burden also hinders the approach from being easily scalable to larger grid sizes.

\subsection{Unit Action Simulation}
UAS works by iterating over unit $i\in\{1,2,...N\}$ controlled by the agent, obtain a sub-action $u_i$, simulate how $u_i$ will play out in the environment and get a new simulated state $s_t^{(i)}$ that is then used to get the action for unit $i+1$. The key here is that only a policy $\pi_{\theta,x,y}(u_t|s_t^{(i)})$ for sub-action space is needed rather than the action space. This results in no wasted logits being generated for empty grid cells as opposed to \emph{GridNet}. The problem is that to acquire the simulated state we need access to the environment's transition probability function $P$ to give us this next state. This effectively makes this approach model-based and it also only reasonable works for deterministic environments, as in a stochastic one, there would be several possible simulated states from each sub-action.


\section{Method}\label{sec:method}
Given an observation $s_t$ of shape $(height \times width \times 27)$ we have developed a scalable policy network that can convert it into $k_t$ logits of shape $(k_t\times 78)$ where $k_t$ is the number units the agent controls. Additionally, we yield a corresponding value estimate $\hat{V}_\theta(s_t)$. A visual overview of the architecture is provided in Appendix \ref{app:net_architecture} and with an explanation of each component being in the following sections. Further, all used hyper-parameters are referenced in Appendix \ref{app:hyper_parameters}.

\subsection{Feature Map}
The first step is to use a feature map $\phi(s_t)$ that converts $s_t$ into $e_t = k_t + l_t + m_t$ individual representations, where $e_t$ is the number of entities, $k_t$ is the number of agent-controlled units, $l_t$ is the number of enemy-controlled units and $m_t$ is the number of neutral units (e.g resource mines). Our implementation of $\phi(s_t)$ is shown in Algorithm \ref{alg:feature_map} and expresses each representation as a one-hot vector of the position (length $hw$) concatenated with the feature vector of the entity (length $27$).

\begin{algorithm}[tb]
\caption{Feature Map $\phi(s_t)$}
\label{alg:feature_map}
\textbf{Input}: $s_t$ with shape $(h\times w \times 27)$\\
\textbf{Output}: $x_t$ with shape $(e_t  \times (hw + 27))$\\
\begin{algorithmic}[1] 
\STATE Let $v_t = \text{reshape}(s_t)$ with shape $(hw \times 27)$
\STATE Let $x_t\leftarrow \mathbf{0}$
\FOR{each entity index $i\in\{1,2,...,e_t\}$}
\STATE $p = \text{position}(v_t,i)$     \hfill\COMMENT{gets index position on grid}
\STATE $x_t[i,\mathbin{:} hw] = \text{onehot}(p)$
\STATE $x_t[i,hw\mathbin{:}] = v_t[p,:]$
\ENDFOR
\STATE \textbf{return} $x_t$
\end{algorithmic}
\end{algorithm}

For example, this means that for the $8 \times 8$ map, $\phi(s_t)$ yields an $(e_t \times (64 + 27)) = (e_t \times 91)$ matrix. Similarly, for the $16 \times 16$ map, we obtain a $(e_t \times (256 + 27) = (e_t \times 283)$ matrix. We can observe that for the larger map, the size of our representation grows significantly. This does not scale well, but fortunately a one-hot encoding can be represented much more densely by using a trainable embedding:

\[ 
\begin{split}
\vect{p}_{embed} = \text{onehot}(p) \matr{W}_{embed}\\
\vect{p}_{embed}\in \mathbb{R}^d, \quad \matr{W}_{embed}\in \mathbb{R}^{(hw\times d)}
\end{split}
\]

This can mitigate the size increase caused by larger maps and in our implementation, we set our embedding weights to be $\matr{W}_{embed}\in \mathbb{R}^{(256\times 64)}$. As a result, the $\phi(s_t)$ generates the same size output of $(e_t \times 91)$ for both the $8 \times 8$ and  $16 \times 16$ map. We should note here, that this size reduction is obviously not infinitely possible and that in particular the representation size should be as least as large as the number of sub-action logits (e.g $91 > 78$). Otherwise it becomes more difficult to obtain meaningful logits.

\subsection{Transformer Net}
With an $x_t = \phi(s_t)$ we can now input this into a transformer encoder. The encoder is composed of 5 stacked, identical transformer layers each of which has 7 attention heads and uses 512 neurons in the dense layer with a ReLU activation \cite{agarap2018deep} (other hyper-parameters detailed in Appendix). In our implementation the entity representations in $x_t$ are ordered as follows by player: player 1 (agent), player 2 (opponent) and neutral units. As we do not use a positional embedding, this does not affect the encoder, but it makes the encoder's output $y_t\in \mathbb{R}^{e_t\times 91}$ easier to process by the actor and critic.

\subsection{Actor \& Critic}
Both the actor and critic networks share this encoder and use $y_t$. The actor takes in all $k_t$ outputs related to the agent's units and then applies a shared single feed-forward layer (with $91\times78+78=7176$ parameters) to each of them. This way the actor is invariant to the number of inputs it receives and manages to yield variable sub-action logits $z_1,...z_{k_t}$.

The critic is a little more complex. As before, we can apply another single feed-forward layer (with $91\times 1+1=92$ parameters), but this leaves us with $e_t$ different value estimates $\hat{V}_{\theta,1}(s_t)...\hat{V}_{\theta,e_t}(s_t)$ that  must somehow be combined into one value $\hat{V}_\theta(s_t)$. A sum naturally comes to mind:

\[  
\hat{V}_\theta(s_t) = \sum_{i=1}^{e_t} \hat{V}_{\theta,i}(s_t)
\]

However, this creates some variance issues. We know that $\hat{V}_{\theta,1}(s_t)...\hat{V}_{\theta,e_t}$ are the values of player units, enemy units and neutral resources. Intuitively, we would expect that a large number of player units means the agent is doing well ($k_t > l_t$), yielding a higher value estimate, whereas a large number of enemy units would yield a lower estimate ($k_t < l_t$). Higher and lower is difficult to model without knowing which value estimate belongs to which player. We could rely on the transformer to provide this information for our critic, but this is an unnecessary burden. Instead, we should try to obtain some more meaningful metrics that the critic can learn to weight on its own:

\[  
\begin{split}
\hat{V}_\theta(s_t) = \sum_{i\in \{p_1,p_2,p_3\}}w_{i,1}\Sigma_{i}+b_{i,1}+w_{i,2}\mu_{i}+b_{i,2} \\
\Sigma_i = \sum_{\text{unit } j \text{ of player } i} \hat{V}_{\theta,j}(s_t), \\
\mu_i = \frac{\Sigma_i}{\text{unit count of player } i} 
\end{split}
\]

Here we calculate 6 metrics: the sum $\Sigma_i$ and mean $\mu_i$ of each player's ($\in \{p_1,p_2,p_3\}$) value estimates  (here $p_3$ refers to the neutral units). We then apply another feed-forward layer (with weights $w_{{p_1},1}...w_{{p_3},2}$ and biases $b_{{p_1},1}...b_{{p_3},2}$) so the critic can choose how to weight each. This allows for the critic to make the value estimates invariant of unit counts and weight units differently based on their owner.

\subsection{Training}
In order to train this model the agent conducts exploratory episodes of up to 2000 steps in 24 parallel instances.  The opposing player in these instances is one of 4 scripted AIs: 

\begin{itemize}
    \item \textit{randomBiasedAI} - chooses random moves for each unit, but is 5 times likelier to harvest resources or attack.
    \item \textit{workerRushAI} - implements a simple rush strategy where it constantly trains workers to attack the closest enemy unit.
    \item \textit{lightRushAI} - implements a simple rush strategy where it builds a barracks and constantly sends light military units to attack.
    \item \textit{coacAI} - a mixed AI that won the 2020 \emph{$\mu$RTS} contest.
\end{itemize}

These AIs and their rankings are publicly available\cite{santiontanon_ai_defs}\cite{microrts_competition_2020} and there are 9 more available in the \emph{Gym-$\mu$RTS} environment, which will be used for evaluation in Section \ref{sec:results}.

As an episode can be quite long, we opted to use \textit{shaped rewards} in order to ensure a more stable and reproducible training \cite{DBLP:journals/corr/abs-1911-08265}. The agent receives a reward whenever it completes any of the following:
\begin{itemize}
    \item $-10$/$0$/$+10$ for losing/drawing/winning, where drawing occurs when the episode step limit is reached with both players still having units.
    \item $+1$ for harvesting a resource from a mine.
    \item $+1$ for attacking an enemy unit.
    \item $+0.2$ for constructing a building.
    \item $+1$ for creating a worker.
    \item $+4$ for producing a combat unit.
\end{itemize}

After collecting ($24\times 256$) steps, we then run the PPO gradient ascent and loss calculation for 4 epochs. Each particular update is conducted in mini-batches of 4. Generally, the opponent selection, reward structure and major training hyper-parameters are kept as similar as possible to \cite{DBLP:journals/corr/abs-2105-13807}, so a more meaningful comparison can be made in the results. 

\section{Results}\label{sec:results}
We trained 2 models for evaluation: one for the $8\times8$ and one for $16\times16$ map. Both models used the same training hyper-parameters with the exception of the $16\times16$ model additionally using the embedding. Accordingly, their trainable parameter counts are very similar: 645470 and 661854 respectively. They were each trained on 2 CPUs and 1 GPU (Tesla V100-SXM2-16GB) with 16GB of RAM for 100 million steps. 

The models were each evaluated for 100 games against each of the 13 scripted AIs available in the $\emph{Gym-$\mu$RTS}$ environment, with a summary of their performance in Table \ref{tab:main_results} and a more detailed breakdown in Appendix \ref{app:win_rates}. Additionally, we benchmark against Huang \emph{et al.}'s results, comparing shaped rewards and also training duration given that similar computational resources were used.

Analysing the results on the $8\times8$ map, the \textit{Transformer Net} has achieved a high win-rate of 91\% within a reasonable training time, although not much lower than any of the models used for the $16\times16$ map. Its shaped reward is significantly lower, but this is to be expected as the $8\times8$ only has half of the harvestable resources of the $16\times16$ map. The proximity to the opponent's base also does not encourage the slow process of building military buildings followed by training high reward military units. Instead, the agent adopts a more refined "worker rush" strategy where it attacks early and quickly only using workers.

\begin{figure*}
     \centering
     \begin{subfigure}[b]{0.327\textwidth}
         \centering
         \includegraphics[width=\textwidth]{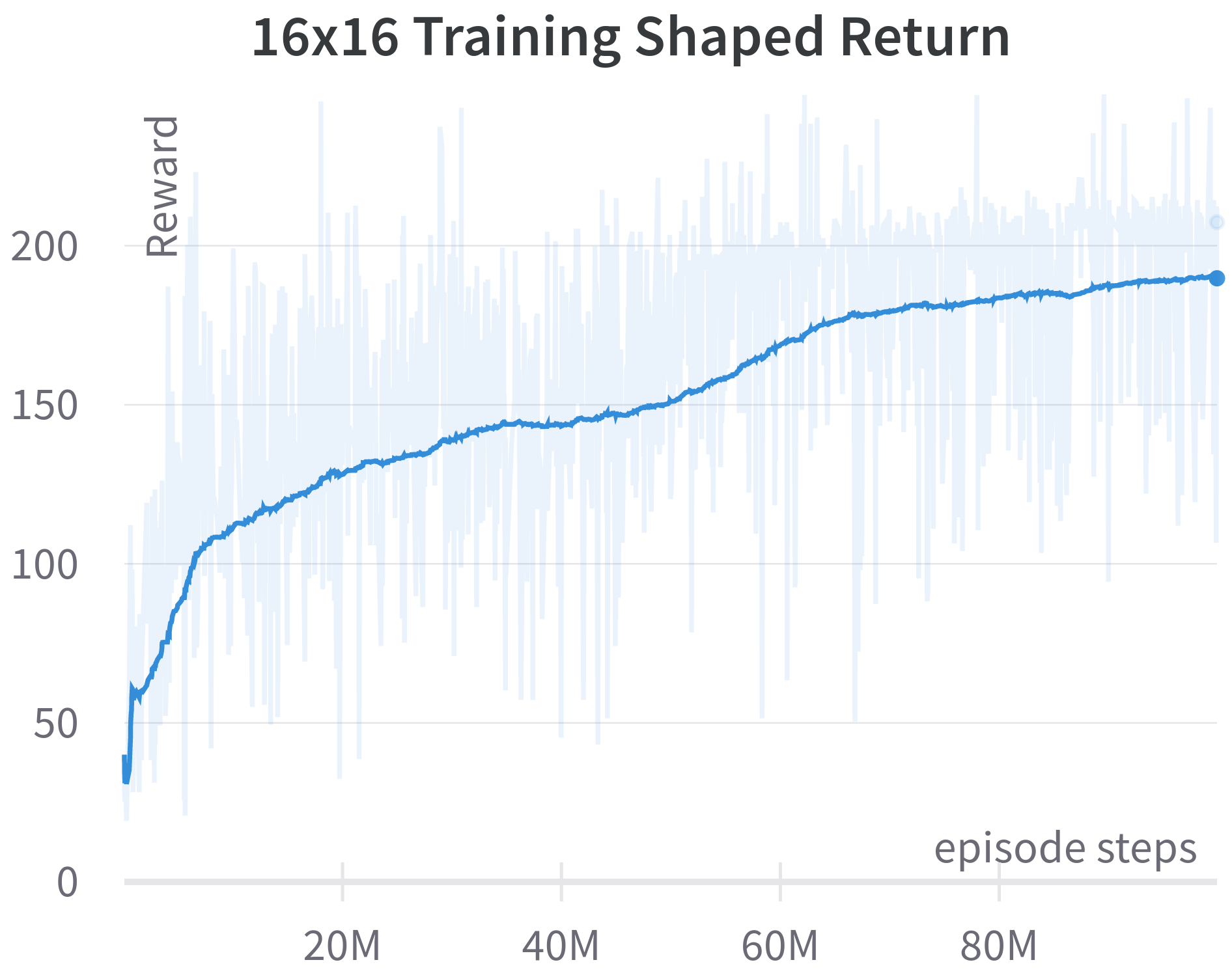}
     \end{subfigure}
     \hfill
     \begin{subfigure}[b]{0.572\textwidth}
         \centering
         \includegraphics[width=\textwidth]{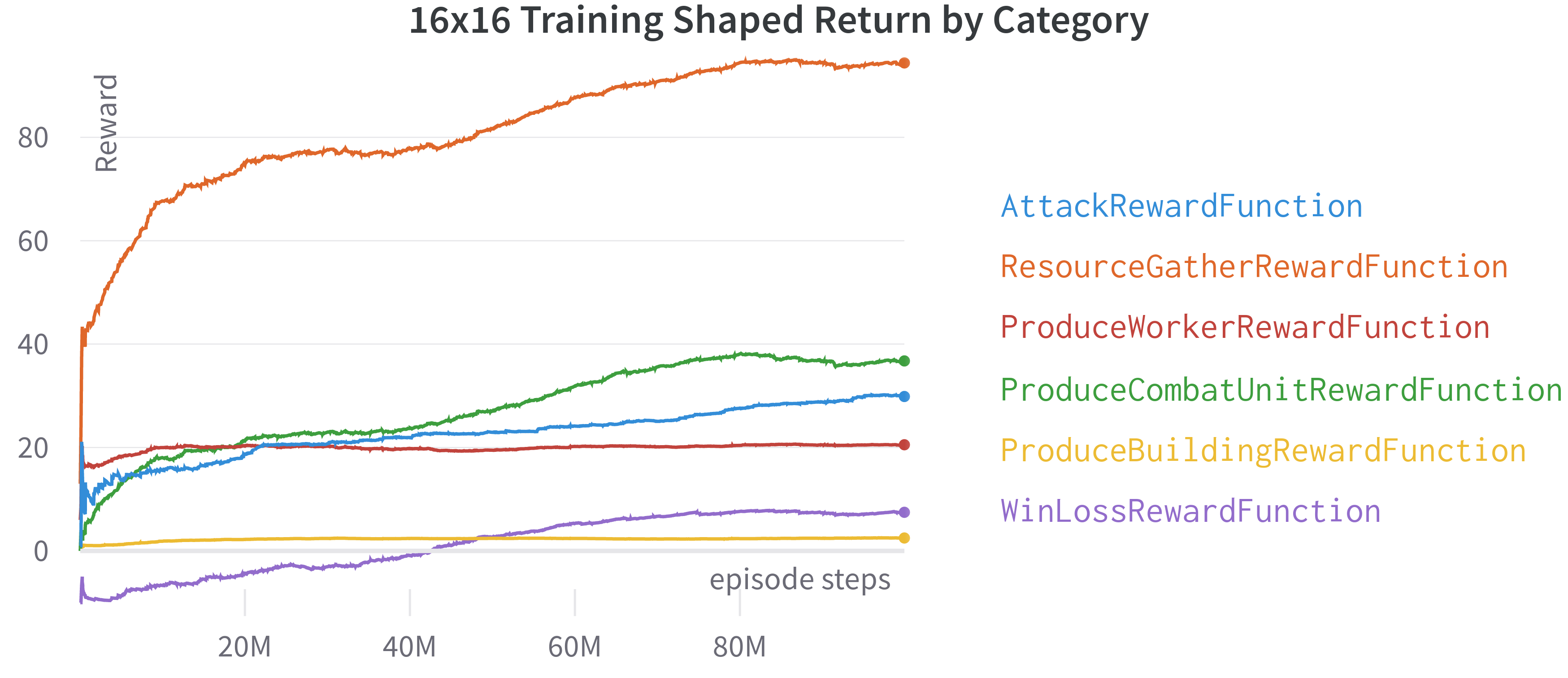}
     \end{subfigure}
    \caption{Shaped return obtained during training of the transformer net on the $16\times16$ map. The chart on the left shows the total shaped return, whereas the one on the right shows a breakdown by reward category. Here we can see that the agent obtained almost half of all rewards from harvesting resources, followed by producing military units. The win/loss reward was much less significant by comparison.}
    \label{fig:16x16_shaped_return}
\end{figure*}

More curious are the results of \textit{Transformer Net w. Embedding} on the $16\times16$ map. It appears to have achieved the highest shaped return of 189.7, but the lowest win-rate of 76\%. The likely cause here is an over-fitting of the shaped rewards. A win/loss reward difference of 20 is simply not significant enough (see Figure \ref{fig:16x16_shaped_return}). An inspection of the evaluation recordings of the agent supports this theory. The agent appears to have learned to cripple the opposing player by defeating all units save for a single survivor. Then it proceeds to artificially prolong the episode by continuing to harvest its own resources and even starts harvesting the opponent's resources. Eventually it finishes off the remaining unit and ends the episode. Unfortunately, this strategy does not always succeed and does not correlate strongly with consistently winning episodes. However, we would argue that this is ultimately an issue of reward design, not the model. Achieving the highest return in a comparatively short training time shows a lot of promise.

\begin{figure*}
     \centering
     \begin{subfigure}[b]{0.327\textwidth}
         \centering
         \includegraphics[width=\textwidth]{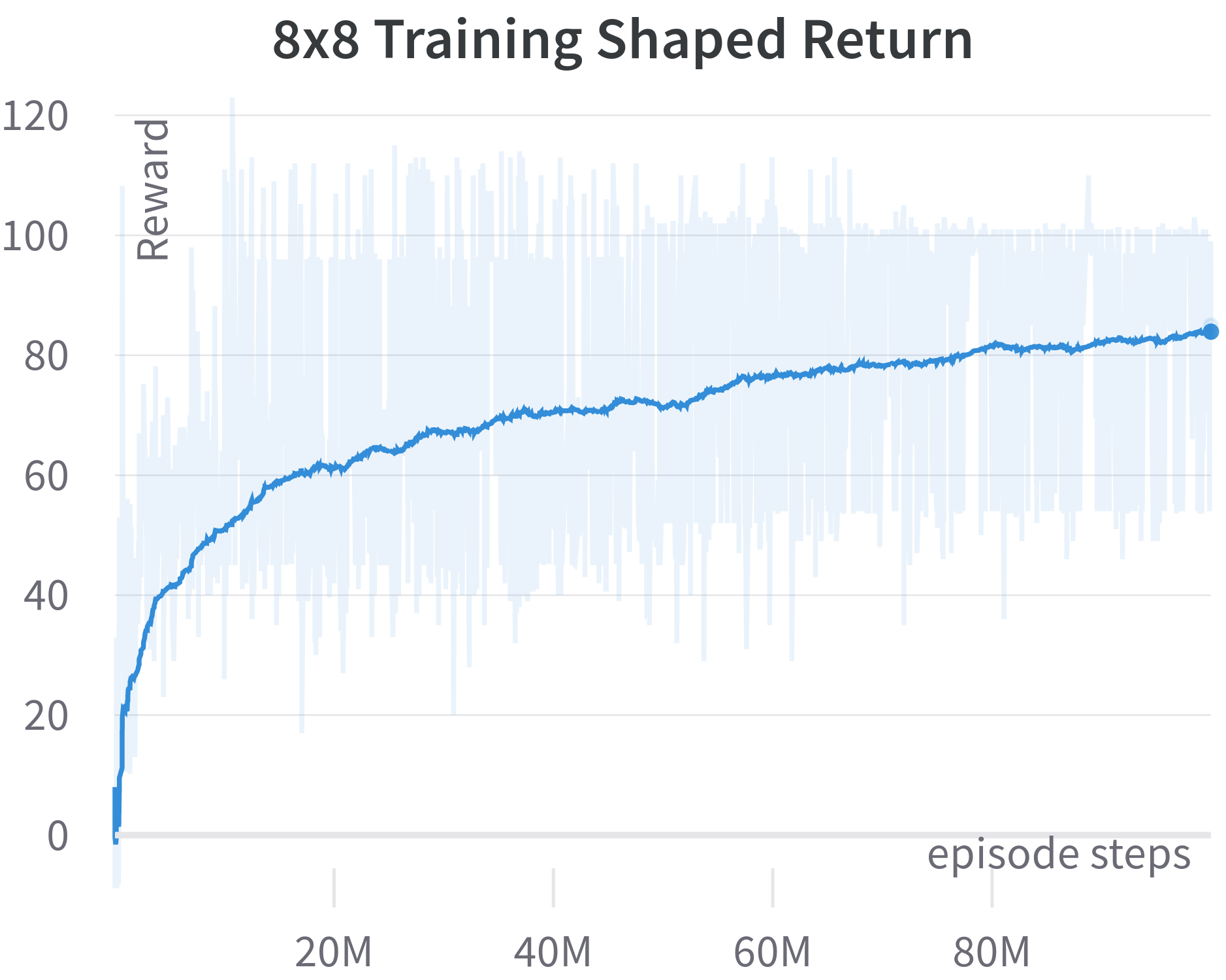}
     \end{subfigure}
     \hfill
     \begin{subfigure}[b]{0.572\textwidth}
         \centering
         \includegraphics[width=\textwidth]{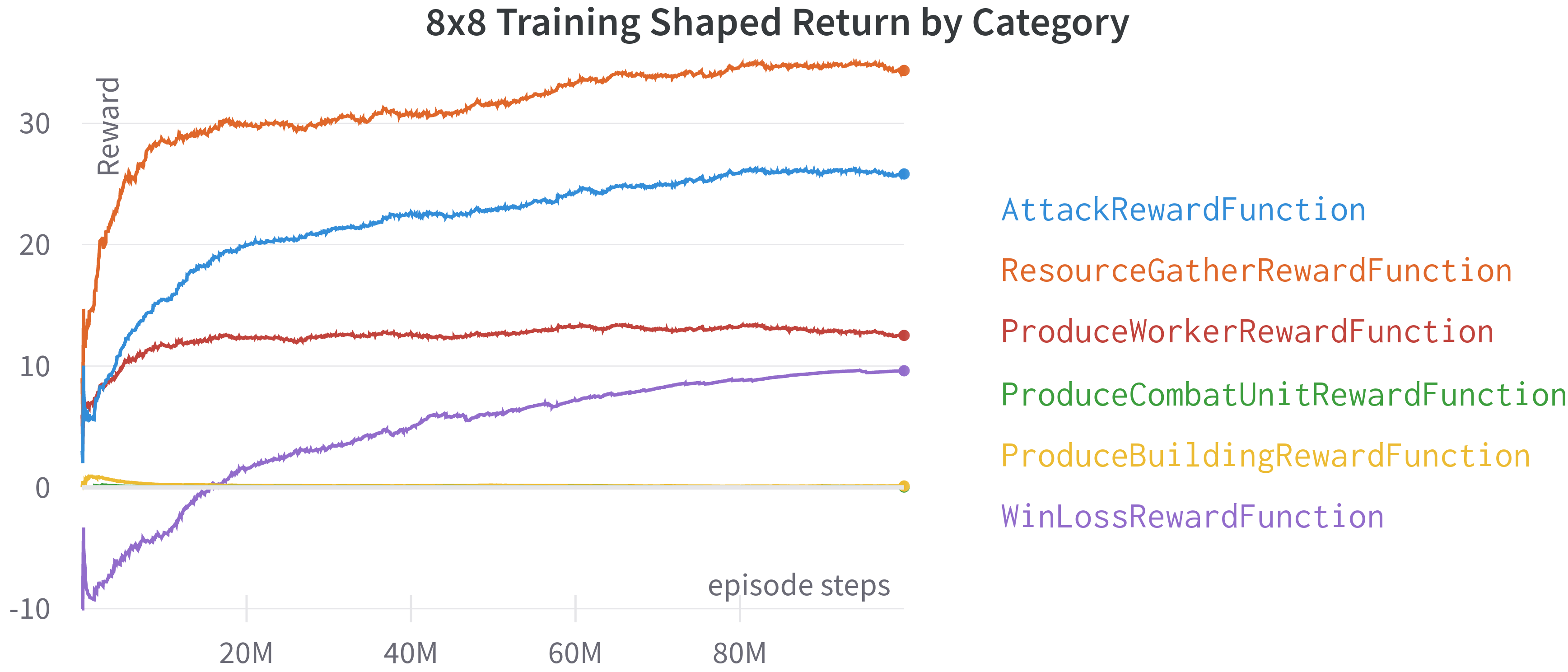}
     \end{subfigure}
    \caption{Shaped return obtained during training of the transformer net on the $8\times8$ map. The chart on the left shows the total shaped return, whereas the one on the right shows a breakdown by reward category. We can observe here that the agent derived the most reward from harvesting resources and attacking. Building barracks and training combat units were not rewarding for it all.}
    \label{fig:8x8_shaped_return}
\end{figure*}

\begin{table*}
\centering
\begin{tabular}{|l|l|l|l|l|} 
\hline
\textbf{Model}               & \textbf{Map Size} & \textbf{Win-rate} & \textbf{Shaped Return} & \textbf{Training Time (hours)}  \\ 
\hline
\textit{GridNet Encoder-Decoder}      & $16\times16$             & 0.89              & 180.8                  & 117.03                          \\ 
\hline
\textit{UAS Impala CNN}                & $16\times16$             & \textbf{0.91}     & 137.6                  & 63.67                           \\ 
\hline
\textit{Transformer Net}              & $8\times8$               & \textbf{0.91}     & 84.9                   & \textbf{55.30}                  \\ 
\hline
\textit{Transformer Net w. Embedding}  & $16\times16$             & 0.76              & \textbf{189.7}         & 58.87                           \\
\hline
\end{tabular}
\caption{Evaluation win-rates, shaped return and training times for currently known best performing RL agents in \emph{Gym-$\mu$RTS}. The results for the \textit{GridNet Encoder-Decoder} and \textit{UAS Impala CNN} are sourced from \cite{DBLP:journals/corr/abs-2105-13807}. It should be noted that most $8\times8$ results are not really comparable to the $16\times16$ as the environments are too different. They are kept in the same table solely for convenience.}
\label{tab:main_results}
\end{table*}

\section{Evaluation \& Future Work}
Overall, we believe this transformer-based policy to be a viable approach for variable action environments such as \emph{Gym-$\mu$RTS}. It has managed to demonstrate a better reward maximisation than \textit{UAS} and faster training time than \textit{GridNet}.

There are of course, a number of open problems. It is not yet known how this transformer policy will work with sparse rewards nor how well it will generalise from self-play. Further, it has yet to be determined how far this can be scaled. Recall that a forward pass through a self-attention layer has a computational complexity of $O(n^2 d)$ \cite{DBLP:journals/corr/VaswaniSPUJGKP17}, where $n$ is the number of inputs and $d$, their length. In \emph{Gym-$\mu$RTS} we recorded a maximum of $n=40$ entities being present on a $16 \times 16$ map at any one time (see Appendix \ref{app:entity_counts}). Making $40^2$ comparisons is quite reasonable, but is this feasible for $n=1000$ or more? This remains to be seen.

\section{Acknowledgements}
I would like to thank Shengyi Huang for his excellent paper\cite{DBLP:journals/corr/abs-2105-13807} and well-documented repository \cite{vwxyzjn}  that I could use as a starting point for this work. Further, I would also like to thank him for taking the time to answer my questions in a friendly and patient manner.

\bibliography{references}

\appendix
\newpage
\onecolumn

\begin{appendices}

\section{Source Code, Trained Models and Recordings}
\label{app:source_code}
We provide additional resources under the following URLs:
\begin{itemize}
    \item source code and trained models: \url{https://github.com/NiklasZ/transformers-for-variable-action-envs}
    \item evaluation game recordings: \url{https://drive.google.com/drive/folders/1I63L0WyEyN0v1rvd7H5SN-27VbdRkvOq?usp=sharing}
    \item training run data and charts: \url{https://wandb.ai/niklasz/public_var_action_transformers}
\end{itemize}

\section{Observation Examples}
\label{app:obs_example}
\begin{figure}[h!]
\centering
\includegraphics[scale= 0.5]{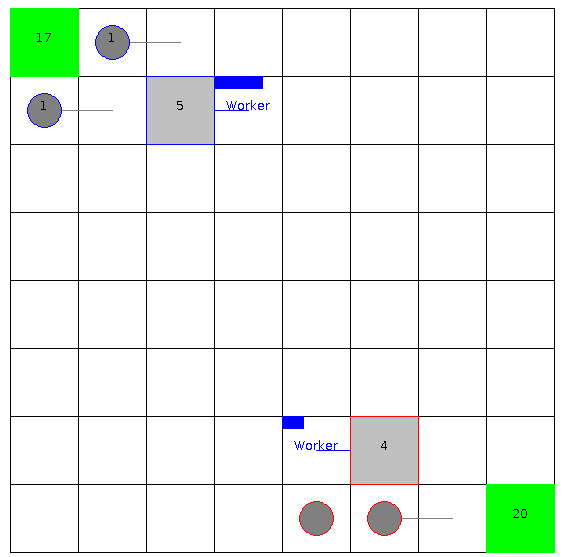}
\caption{The above depicts 2 tiles of resources in green, and 1 base and 2 workers for each player. Each base is currently constructing a worker. The blue workers have a "1" on their sprite as they are each carrying a resource.}
\label{fig:obs_example_board}
\end{figure}

Consider the example board in Figure \ref{fig:obs_example_board}. We can express the top-left resource cell using the following encoding:

\[  
[1,0,0,0,0],[0,0,0,0,1],[0,1,0],[0,1,0,0,0,0,0,0],[1,0,0,0,0,0]
\]

Where the first vector tells us it has $0$ HP, the second tells us it has $\geq4$ resources, the 3rd tells us it is owned by the neutral player, the 4th tells us it is a resource and the last one tells us it is presently doing nothing. Similarly, we can express the worker to the right of it using:

\[  
[0,1,0,0,0],[0,1,0,0,0],[1,0,0],[0,0,0,0,1,0,0,0],[0,1,0,0,0,0]
\]

In this example we see the worker has $1$ HP and is carrying $1$ resource. It belongs to player 1 and is a worker unit. Its current action (indicated by the line pointing out of it) is to move.

\section{Agent Neural Network Architecture}
\label{app:net_architecture}
\begin{landscape}
\begin{figure}
\centering
\includegraphics[scale= 0.120]{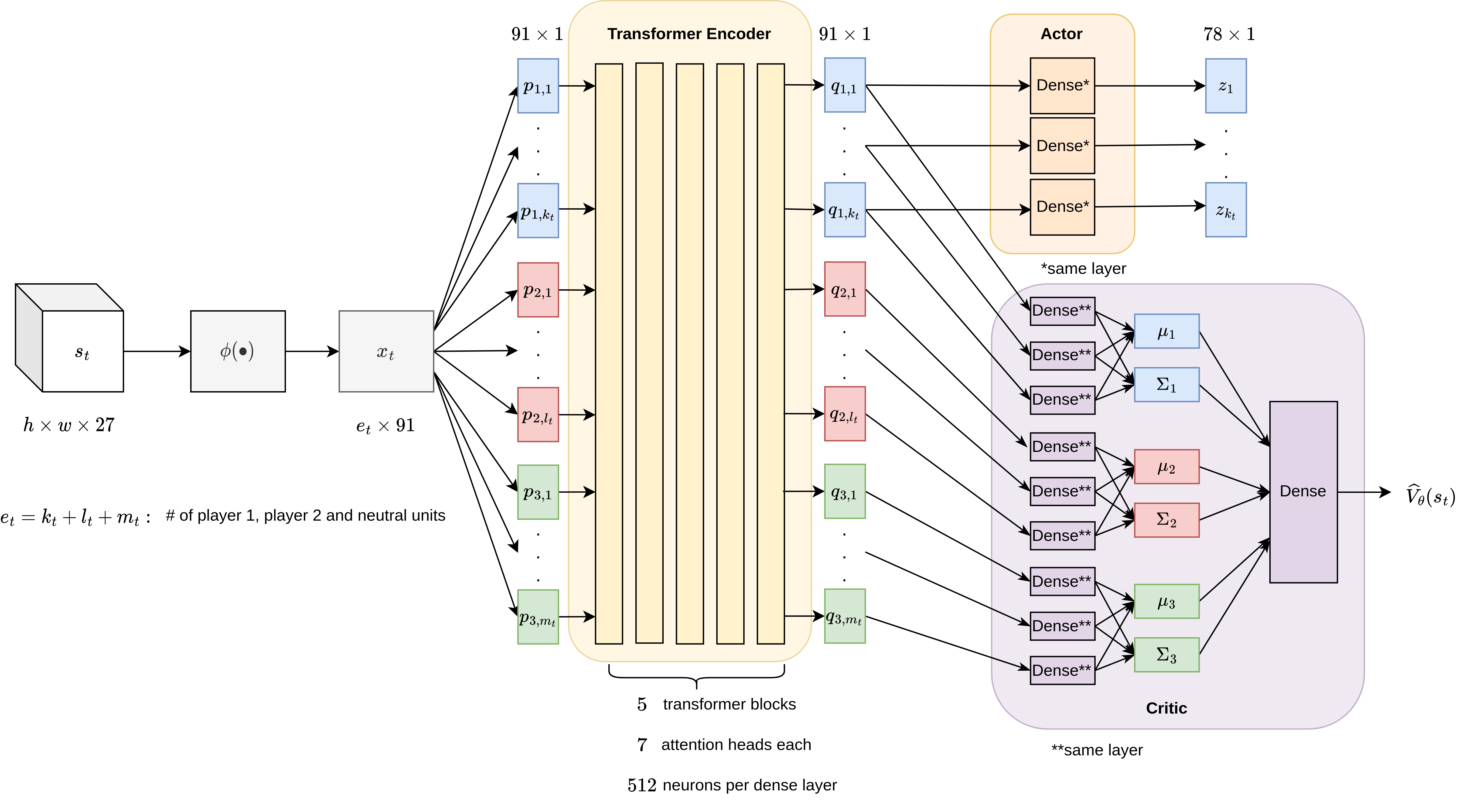}
\caption{Neural Network Architecture of the Actor and Critic. We first apply the feature map $\phi(s_t)$ and then order the resulting vectors by unit ownership: player 1, player 2, player 3 (neutral entities like resources). Next we feed it through the transformer encoder layers. The actor then takes the outputs from player 1 and creates action component logits $z_1...z_{k_t}$, whereas the critic considers the outputs from all players. From them, it calculates respective mean and sum values which it then weights into a final value prediction $\hat{V}_\theta(s_t)$. Lastly, note that in the actual implementation we also include a batch dimension $B$, meaning the real input has shape $B \times h \times w\times 27$ and the actor output will be $B$ lists of logits $z_1...z_{k^b_t}$ where $k^b_t$ will differ for each list $b\in B$.}
\end{figure}
\end{landscape}

\section{Agent Win-Rates}
\label{app:win_rates}

\begin{figure}[h!]
\centering
\includegraphics[scale= 0.1]{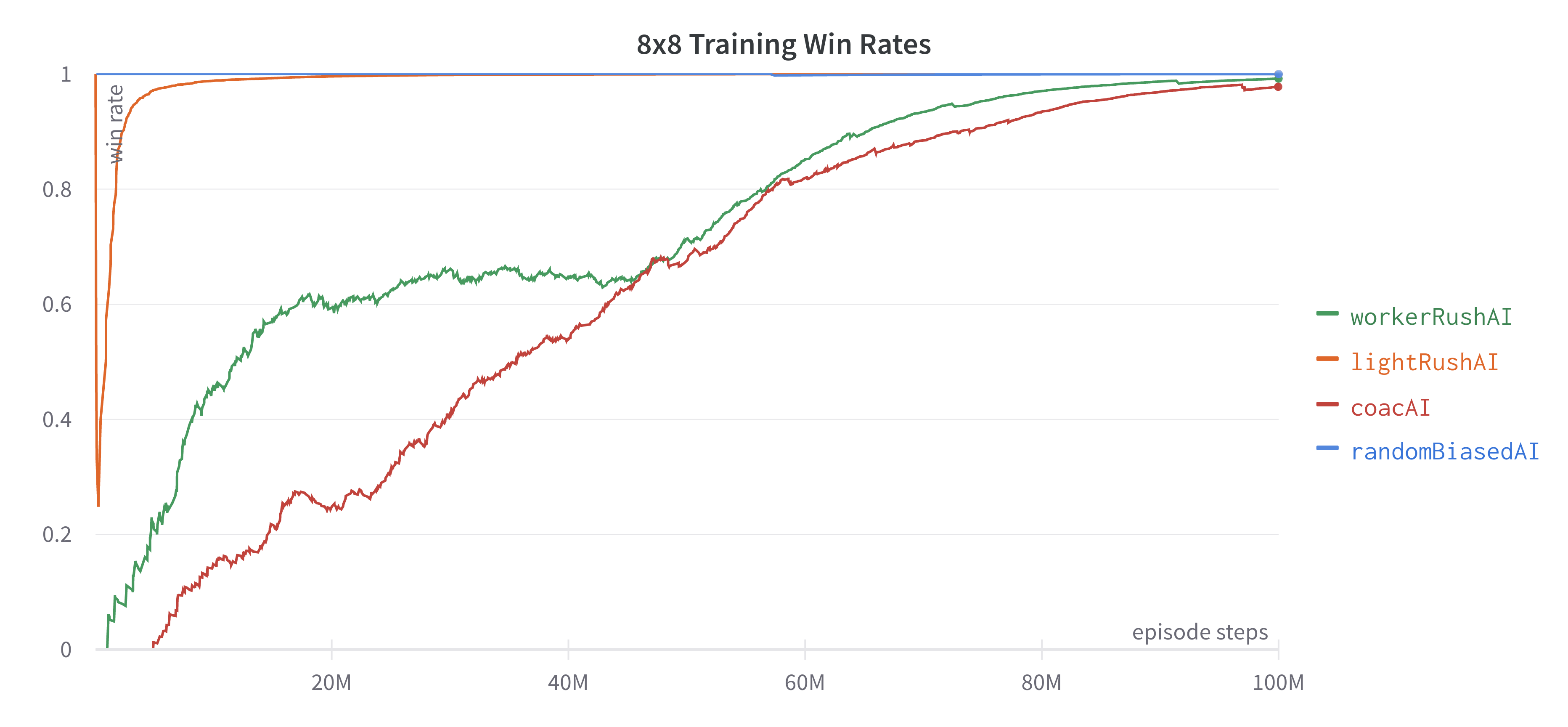}
\caption{Win-rate of the $8\times8$ agent during training, smoothed via exponentially moving average. It took the longest to train against the \textit{workerRushAI} and \textit{coacAI}.}
\label{fig:win}
\end{figure}

\begin{figure}[h!]
\centering
\includegraphics[scale= 0.3]{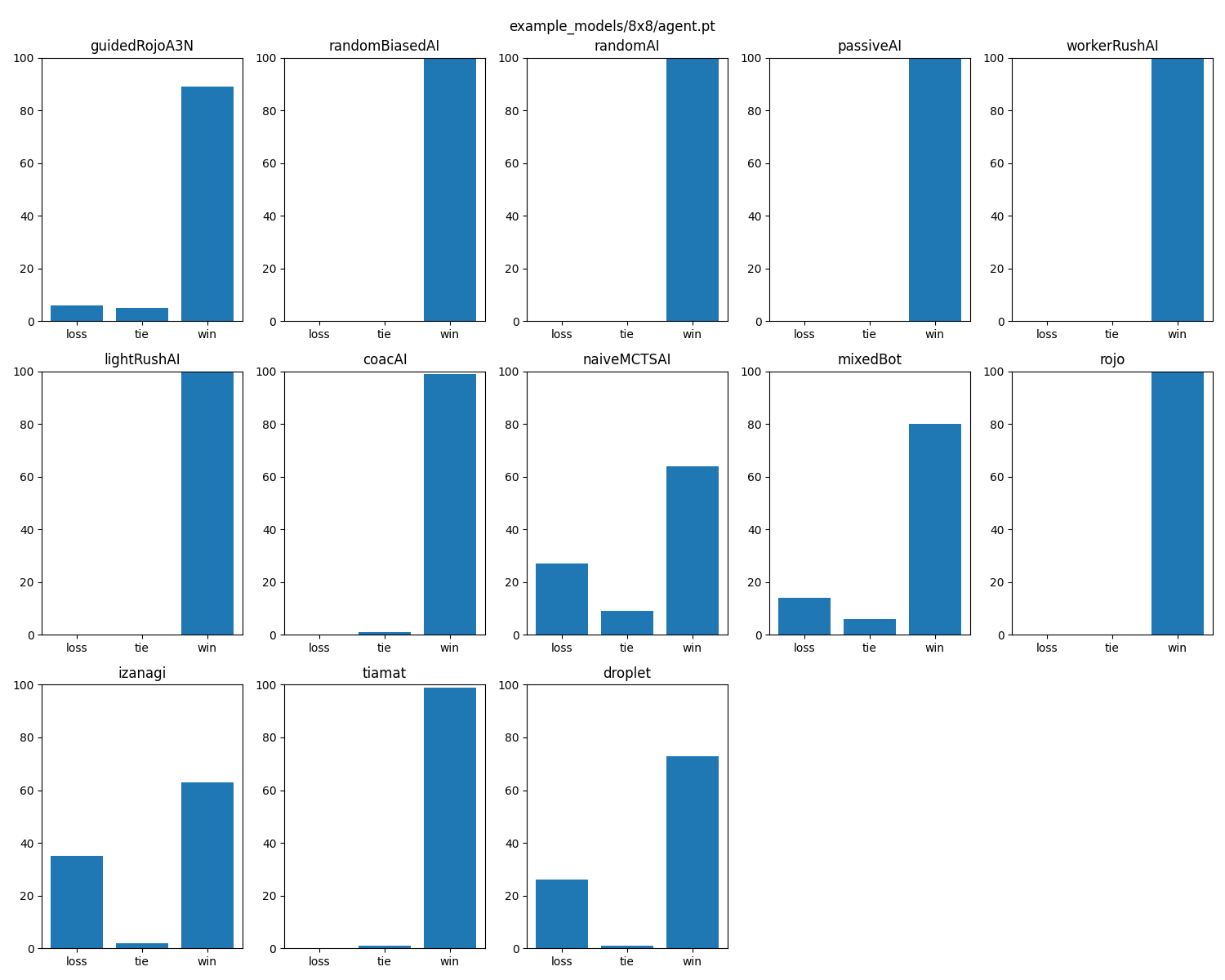}
\caption{Match stats of the $8\times8$ agent during evaluation.  Ties occur when neither AI manages to completely eliminate the opponent within the episode step limit $T=2000$.}
\label{fig:win_rates_by_opponent}
\end{figure}

\begin{figure}[h!]
\centering
\includegraphics[scale= 0.1]{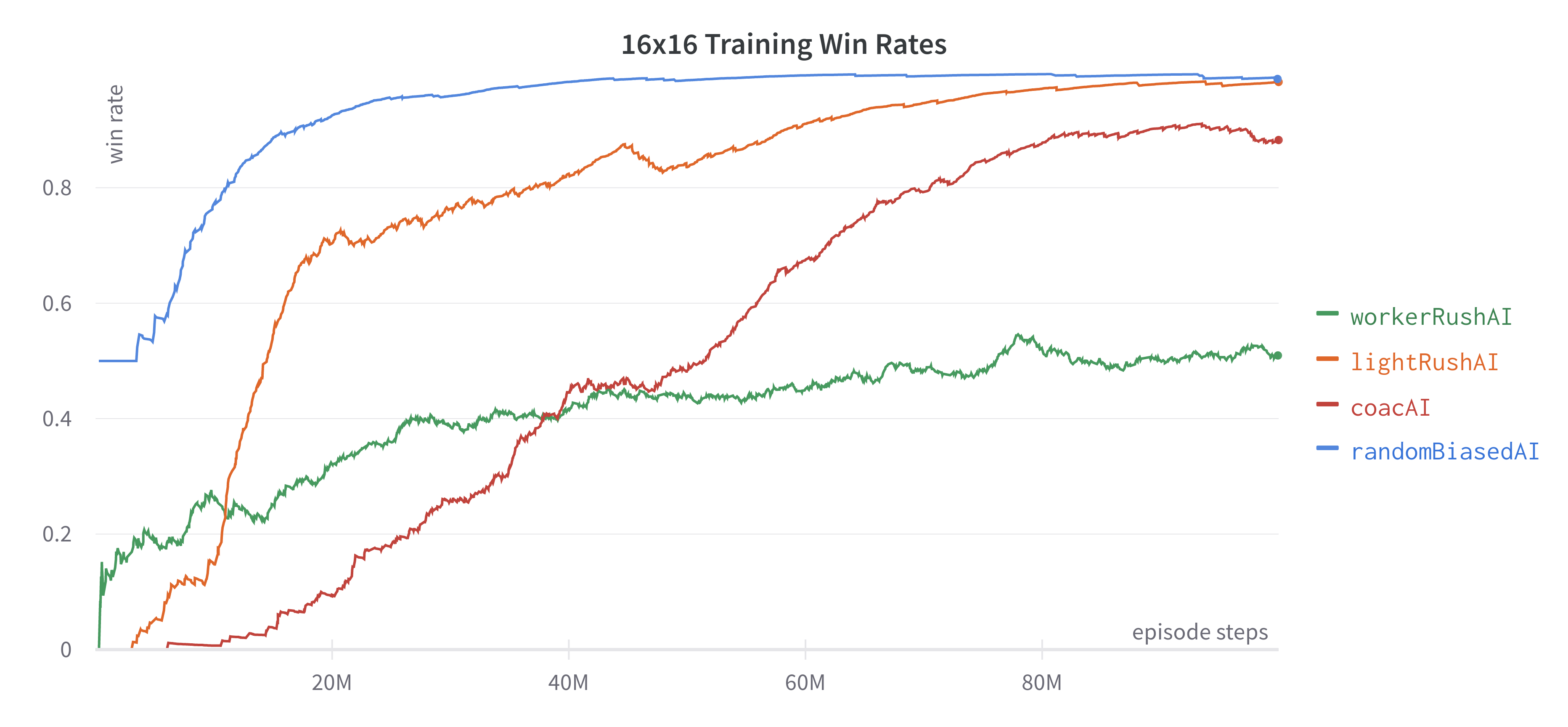}
\caption{Win-rate of the $16\times16$ agent during training, smoothed via exponentially moving average. We can observe that the agent managed to win most of the time against its training opponents, but never achieved a decisive win-rate. Towards the end, the win-rate even degraded, likely because the agent was not sufficiently rewarded/punished for winning/losing.}
\label{fig:win}
\end{figure}

\begin{figure}[h!]
\centering
\includegraphics[scale= 0.3]{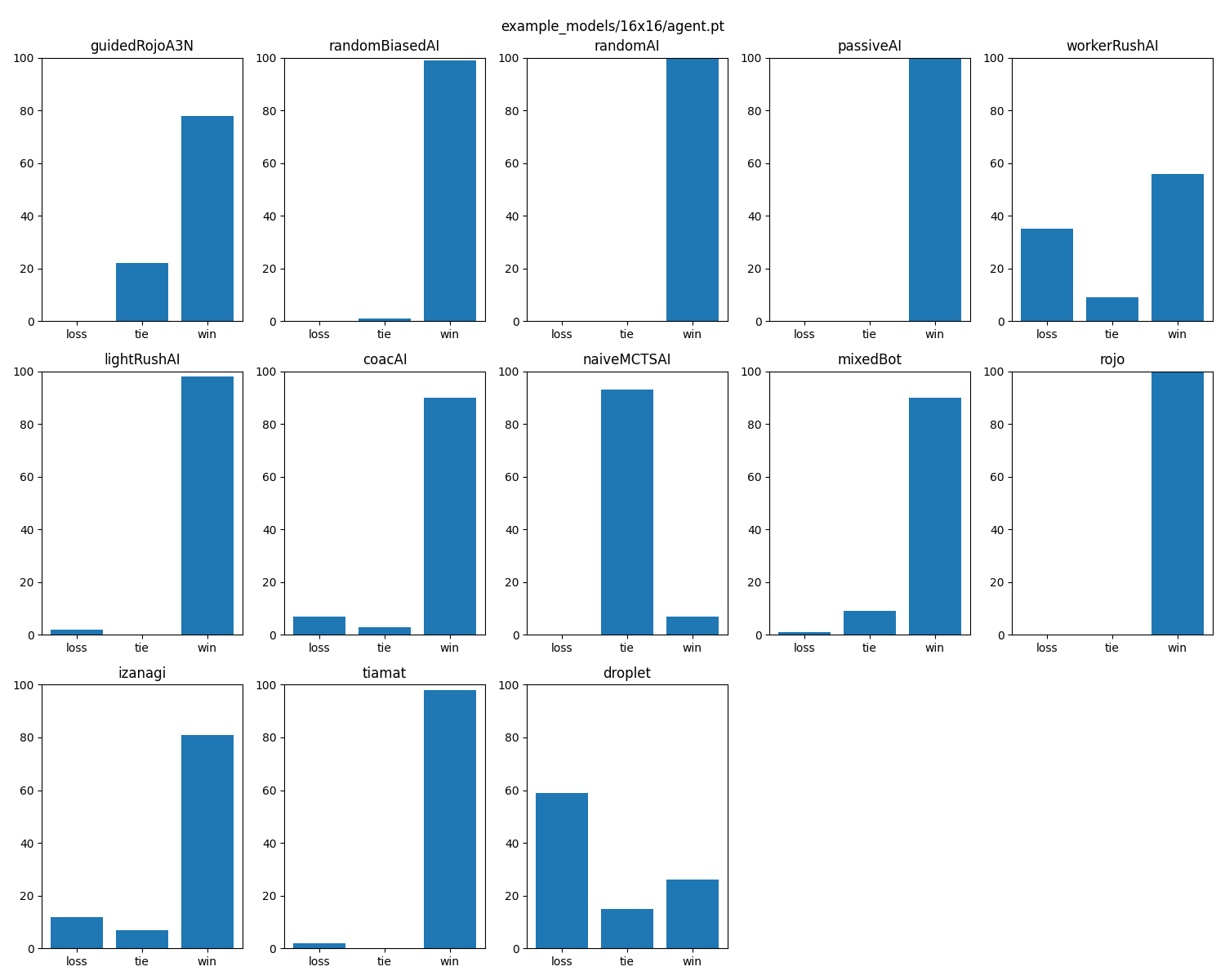}
\caption{Match stats of the $16\times16$ agent during evaluation.  Ties occur when neither AI manages to completely eliminate the opponent within the episode step limit $T=2000$.}
\label{fig:win_rates_by_opponent}
\end{figure}

\section{Agent Hyper-parameters}
\label{app:hyper_parameters}

\begin{table}[h!]
\centering
\begin{tabular}{|>{\hspace{0pt}}m{0.2\linewidth}|>{\hspace{0pt}}m{0.261\linewidth}|>{\hspace{0pt}}m{0.478\linewidth}|} 
\hline
\textbf{Hyper-parameter Name}    & \textbf{Setting}                                                                                                                                      & \textbf{Notes}                                                                            \\ 
\hline
Transformer Layers               & 5                                                                                                                                                     & Chosen to be the best efficacy/efficiency trade-off from ranges of 3 to 8.                                               \\ 
\hline
Transformer feed-forward neurons & 512                                                                                                                                                   & \textcolor[rgb]{0.141,0.161,0.184}{-}                                                     \\ 
\hline
Transformer attention heads      & 7                                                                                                                                                     & \textcolor[rgb]{0.141,0.161,0.184}{Chosen mostly as it fits the input size $91|7$}         \\ 
\hline
Transformer activation function  & ReLU                                                                                                                                                  & -                                                                   \\ 
\hline
Transformer Dropout              & 0.1                                                                                                                                                   & -                                                                   \\ 
\hline
Weight Initialisation          & $\mathcal{N}(0,2)$                                                                                                                                    & -                                                                                         \\ 
\hline
Bias Initialisation              & 0                                                                                                                                                     & -                                                                                         \\ 
\hline
Optimiser                        & Adam with:\par{}$\alpha = 2.5\mathrm{e}{-4}$\par{}$\epsilon = 1\mathrm{e}{-5}$\par{}$\beta_1=0.9$\par{}$\beta_2=0.999$\par{}with a linear decay to 0. & -                                                                                         \\ 
\hline
Max Training Steps               & 100 Million                                                                                                                                           & This refers to the number of episode steps that have been used for gradient calculation.  \\ 
\hline
Number of exploration steps      & 256                                                                                                                                                   & How many steps to traverse per environment before running PPO.                            \\ 
\hline
Parallel bot environments        & 24                                                                                                                                                    & Number of simultaneous environments for exploration.                                      \\ 
\hline
Minibatch size                   & 4                                                                                                                                                     & -                                                                                         \\ 
\hline
Return discount factor           & $\gamma=0.99$                                                                                                                                         & -                                                                                         \\ 
\hline
Generalised Advantage Estimate   & $\lambda=0.95$                                                                                                                                        & -                                                                                         \\ 
\hline
Entropy coefficient              & $c=0.01$                                                                                                                                              & Controls weighting of entropy loss.                                                       \\ 
\hline
Value Function coefficient       & $v_{coef}=0.5$                                                                                                                                        & Controls weighting of value function loss.                                                \\ 
\hline
Gradient clipping coefficient    & $\epsilon=0.1$                                                                                                                                        & PPO clipping coefficient                                                                  \\
\hline
\end{tabular}
\caption{Training hyper-parameters}
\end{table}

\section{Entity Distribution Counts}
\label{app:entity_counts}

\begin{figure}[h!]
\centering
\includegraphics[scale= 0.8]{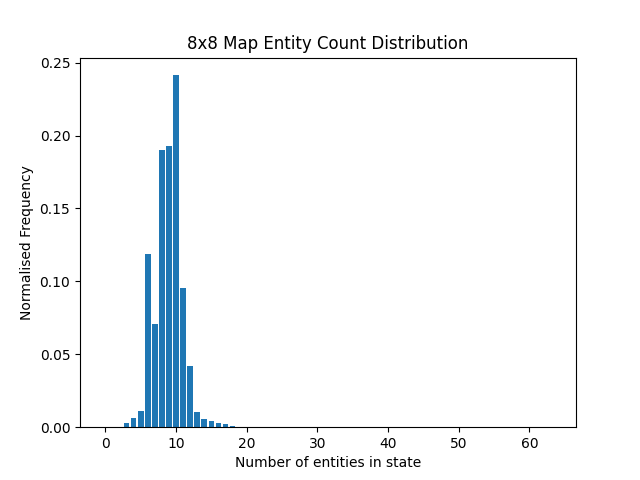}
\caption{Distribution of entities counted in every state during evaluation of the $8 \times 8$ transformer net. On average around 12 entities are present, corresponding to roughly $19\%$ of the cells in the grid.}
\end{figure}

\begin{figure}[h!]
\centering
\includegraphics[scale= 0.8]{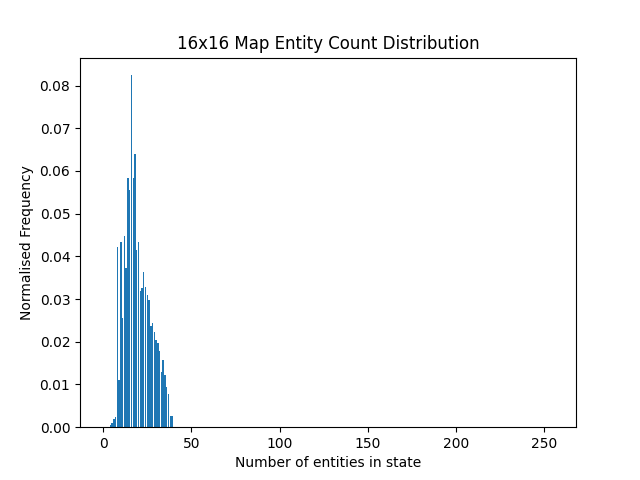}
\caption{Distribution of entities counted in every state during evaluation of the $16 \times 16$ transformer net. On average around 20 entities are present, corresponding to roughly $8\%$ of the cells in the grid.}
\end{figure}

\end{appendices}

\end{document}